\documentclass{article}

\usepackage{arxiv}

\usepackage[utf8]{inputenc} 
\usepackage[T1]{fontenc}    
\usepackage{hyperref}       
\usepackage{url}            
\usepackage{booktabs}       
\usepackage{amsfonts}       
\usepackage{nicefrac}       
\usepackage{microtype}      
\usepackage{lipsum}
\usepackage{amsmath}
\usepackage{amssymb}
\usepackage{graphicx}
\title{Looking Ahead: Anticipating Pedestrians Crossing with Future Frames Prediction}

\author{
  Mohamed Chaabane\thanks{These two authors contributed equally} \\
  Department of Computer Science\\
  Colorado State University\\
  \texttt{chaabane@colostate.edu} \\
   \And
 Ameni Trabelsi\,$^\ast$ \\
  Department of Computer Science\\
  Colorado State University\\
  \texttt{ameni.trabelsi@colostate.edu} \\
   \And
 Nathaniel Blanchard \\
  Department of Computer Science\\
  Colorado State University\\
  \texttt{nathaniel.blanchard@colostate.edu} \\
   \And
 Ross Beveridge \\
  Department of Computer Science\\
  Colorado State University\\
  \texttt{ross.beveridge@colostate.edu} 
}

\begin{document}
\maketitle

\begin{abstract}

In this paper, we present an end-to-end future-prediction model that focuses on pedestrian safety. Specifically, our model uses previous video frames, recorded from the perspective of the vehicle, to predict if a pedestrian will cross in front of the vehicle. The long term goal of this work is to design a fully autonomous system that acts and reacts as a defensive human driver would --- predicting future events and reacting to mitigate risk. We focus on pedestrian-vehicle interactions because of the high risk of harm to the pedestrian if their actions are miss-predicted. 
Our end-to-end model consists of two stages: the first stage is an encoder/decoder network that learns to predict future video frames.
The second stage is a deep spatio-temporal network that utilizes the predicted frames of the first stage to predict the pedestrian's future action. 
Our system achieves state-of-the-art accuracy on pedestrian behavior prediction and future frames prediction on the Joint Attention for Autonomous Driving (JAAD) dataset. \footnote{Accepted as an oral presentation in WACV-20}
\end{abstract}

\section{Introduction}\label{introduction}

For the last decade, researchers and companies alike have been striving to achieve Level 5 autonomy for self-driving vehicles (i.e., autonomous operation with no human intervention) \cite{badue2019self}. One benefit of widespread Level 5 autonomy would be a reduction in vehicular accidents caused by human error, which kills around $1.35$ million people a year \cite{world2018global}. 
The feasibility of Level 5 has been buoyed by research overcoming several milestones, such as autonomous driving on highways \cite{dickmanns1987curvature}, rough terrains \cite{thrun2006stanley}, and urban environments \cite{urmson2008autonomous}. These breakthroughs have led several companies to invest in consumer-grade autonomous vehicles. 
\begin{figure}[ht]\centering
    \includegraphics[scale = 0.8]{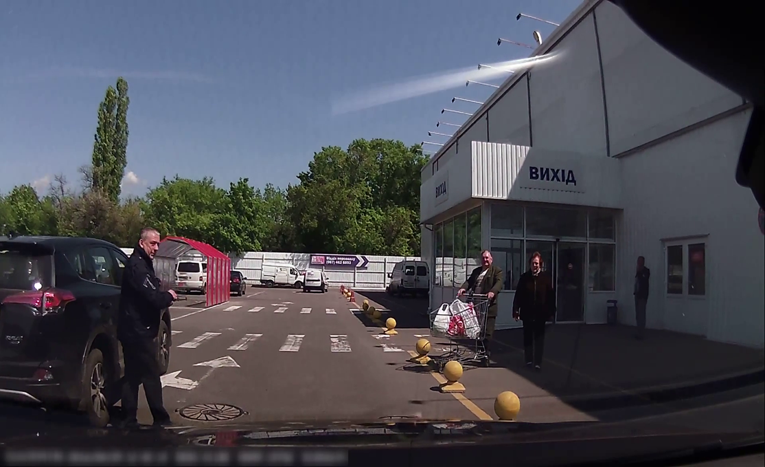}
    \caption{Will the pedestrian step in the vehicle's path? Human drivers and pedestrians interact through nonverbal, physical  communication. For example, a pedestrian might make eye contact with a driver. From this communication, the driver and pedestrians predict each other's actions. The driver may anticipate the pedestrian will attempt to cross the road, and slow down to facilitate. Fully autonomous vehicles are expected to identify the signals pedestrians exude, predict potential actions, and react appropriately. This work presents a pedestrian prediction model that embodies these principles.}
    \label{fig:teaser}
\end{figure}
Nonetheless, there are still many hurdles left to overcome before Level 5 autonomy is reached. One issue is that these vehicles are not replicating the behavior of human drivers --- specifically, the ways human drivers communicate with each other, and surrounding agents \cite{mahadevan2018communicating,sucha2017pedestrian}. 
For example, pedestrians use nonverbal cues such as eye gaze or hand gestures to communicate their crossing intent to human drivers. Conversely, in autonomous vehicles, any misunderstanding of pedestrians gestures or misprediction of their intents will more likely cause traffic accidents. In Figure \ref{fig:teaser}, we present an uncertain situation where a pedestrian may or may not cross in front of the vehicle. Incorrectly predicting that the pedestrian will not cross would likely lead to pedestrian injury. 


This work focuses on the understanding and anticipation of pedestrian action as a step toward an autonomous vehicle capable of understanding such nonverbal communication. To this end, we employ the Joint Attention for Autonomous Driving (JAAD) dataset \cite{rasouli2017they}, which is specifically designed to capture the visual and behavioral complexity between pedestrians and drivers from the perspective of the driver. JAAD consists of a multitude of complexities and conditions (e.g., weather, location, etc.,) that allow us to thoroughly test the robustness of our model. 
The model itself operates in two stages: first, the network takes $N$ video frames (the past) and predicts the next $N$ video frames (the future). Specifically, the input frames are encoded into features using spatio-temporal 3D-Convolution layers. The features are then decoded into predictions of the next future frames using depth-wise separable convolutional LSTM layers. The second stage of the model utilizes the predicted $N$ future frames to classify if the pedestrian will step in front of the vehicle. This stage processes the predicted frames with a supervised action network to predict pedestrians crossing actions. The full model is trained in an end-to-end fashion to minimize loss on both future frames prediction and pedestrian crossing prediction.
Our model is able to achieve state-of-the-art accuracy on both future frames prediction and pedestrian future action prediction. An ablation study of the model's components shows our model is capable of capturing the key aspects of both egocentric and external movement required for accurate pedestrian prediction. The robustness of the future frames prediction component allows the action classification component to achieve state-of-the-art accurately, further indicating the predicted frames have the level of detail required to make such predictions. 

In the future, these models may prove useful to the understanding and prediction of other environmental variables, such as predicting the behavior of other vehicles. The model may also be useful for predicting pedestrian intent, rather than action, something that can be inferred from additional labels provided by the JAAD dataset. For now, we focus on anticipating one of the most dangerous circumstances: a pedestrian stepping in front of a vehicle. 

In summary, this paper makes the following contributions:
\begin{itemize}
    \item We present a future video frames prediction encoder/decoder network that operates in unsupervised manner to predict $N$ future frames of a video using $N$ initial frames.
    \item We propose an end-to-end model that predicts the future video frames and uses the predicted frames as input for supervised action recognition network to predict when pedestrians will step in front of the vehicle. 
    \item We achieve state-of-the-art performance on both future frames prediction and predicting pedestrian future crossing action on JAAD dataset. 
    \item We conduct a thorough ablation study that shows the model components are robust, efficient, and effective across a multitude of weather conditions, locations, and other variables.

\end{itemize}

\section{Related Work}\label{Related_Work}
Our work is related to two avenues of previous work: future video frames prediction and pedestrian future action prediction. 

\textbf{Future Video Frames Prediction:}
In recent years, Recurrent Neural Networks (RNN) have been widely used in future video frames prediction \cite{srivastava2015unsupervised, wang2018predrnn++}. Much of this work focuses on predicting one frame into the future, with an option to modify the network for long range prediction by taking in the predicted frame as input. Some examples of this include \cite{finn2016unsupervised,lotter_deep_2017,srivastava2015unsupervised,sutskever2014sequence,wang2018predrnn++}. Our work differentiates itself by focusing on an architecture explicitly designed to predict many frames into the future.  

Other studies have applied stochastic variational methods to future frames prediction, such as \cite{babaeizadeh2017stochastic} and \cite{denton2018stochastic}.
Several others have worked with generative adversarial networks \cite{denton2017unsupervised,tulyakov2018mocogan,vondrick2016generating}.  These studies focus on the sharpness of the
generated video frames, treating it as
a major characteristic to distinguish between real and fake video frames. However, models from these techniques are often unstable and difficult to properly evaluate. We focus the training and evaluation of our model on traditional dataset-driven performance, which allows us to accurately understand the applications and limitations of our results. 

Recently, Gujjar et al. \cite{gujjar2019classifying} studied the prediction of urban pedestrian actions by predicting future frames of traffic scene videos. Of these related works, this study's approach is the most similar to ours. The key difference is that our network provides better spatio-temporal representations for next frames prediction. This distinction is due to our network's deeper nature and the use of residual connections, which enable it to extract more complex features without falling into vanishing gradient problem. Our network also has a reduced running time and improved performance, thanks to the use of depth-wise separable convLSTMs rather than standard convLSTMs (see Section \ref{ablation}). 
Furthermore, our network uses lateral connections to reduce the blur in future frames, something  especially important for long term prediction.

\textbf{Pedestrian Future Action Prediction:}
In this work, we are mainly concerned with modeling pedestrian future action in the context of autonomous driving cars. Many existing approaches are based on Hidden Markov Model (HMM) where the pedestrian's intent is represented in the hidden state \cite{kelley2008understanding,zhu1991hidden}. These approaches have been extended to combining the motion models of all the intentions together into a single Mixed Observability Markov Decision Process (MOMDP), which is a structured variant of the more common Partially Observable Markov Decision Process (POMDP) \cite{bandyopadhyay2013intention}. Although models utilizing the Markovian process are known for their fast adaptability, their assumption can be restrictive due to insufficient prior conditioning. Thus, the main limitation of the presented methods is their lack of memory. Our approach overcomes this limitation by using RNN, known for its good memory retention qualities, in our model, thus extending its long-term memory.

Other approaches for prediction of time series assume they are samples from a process generated by a linear process driven by a white, zero-mean, Gaussian input \cite{karasev2016intent,schneider2013pedestrian}. Although they can be more accurate, Gaussian processes have shown to be slower than Markov models since they use the entire observed trajectory to predict the future state \cite{ellis2009modelling}. Switching linear dynamical models, applied in constrained environments, were introduced as extensions to these models \cite{karasev2016intent,kooij2014context}. Kooij et al. proposes a Dynamic Bayesian Network for pedestrian path prediction which incorporates environment parameters such as pedestrian situational awareness and head orientation with a Switching Linear Dynamical System to predict changes in the dynamics of pedestrians \cite{kooij2014context}.
These motion models require accurate and precise segmentation and tracking of pedestrians to be efficient. Such assumption can be challenging due to the difficulty of extracting reliable image features for segmentation and tracking \cite{volz2016data}.

Consequently, many approaches, including ours, study pedestrian activity models that are extracted directly from the image space of the captured scenes. 
Hasan et al. \cite{hasan2016learning} treat the prediction of adverse pedestrian actions as an anomaly detection problem. They built a fully convolutional autoencoder to learn the local features followed by a classifier to capture the regularities. 
Rasouli et al. \cite{rasouli2017they} extract context features extracted from input frames using AlexNet \cite{krizhevsky2012imagenet} and train a linear SVM model to predict future crossing action of pedestrians on JAAD dataset. These approaches are limited because they focus only on spatial appearances, ignoring the temporal coherence in long-term motions. To solve this issue, Gujjar et al. \cite{gujjar2019classifying} processes the crossing actions classification by feeding the predicted frames of their future frame prediction network to a C3D based network \cite{tran2014c3d} which takes into account the temporal dynamics in addition to the spatial appearances. As mentioned before, this work is similar to ours, but varies in the training strategy employed, the experimental study of network components, and our network's higher performance with shorter running time (see Section \ref{results_FFP}). 

\section{Methods}\label{Methods}
Here, we detail the end-to-end model (Section \ref{arch}) and enumerate experimental details (Section \ref{experiments}), including an overview of the dataset (Section \ref{data}) and model search procedure (Section \ref{model_search}).

\subsection{Architecture}\label{arch}

Our end-to-end model consisted of two stages: the first stage was an unsupervised encoder/decoder network that generated predicted future video frames. The second stage was a deep spatio-temporal action recognition network that utilized the generated video frames to predict pedestrian action --- specifically, if the pedestrian would cross in front of the vehicle. 

\subsubsection{Future Frames Prediction Component}\label{FFP}

\begin{figure*}[ht]
\begin{center}
   \includegraphics[scale = 0.37]{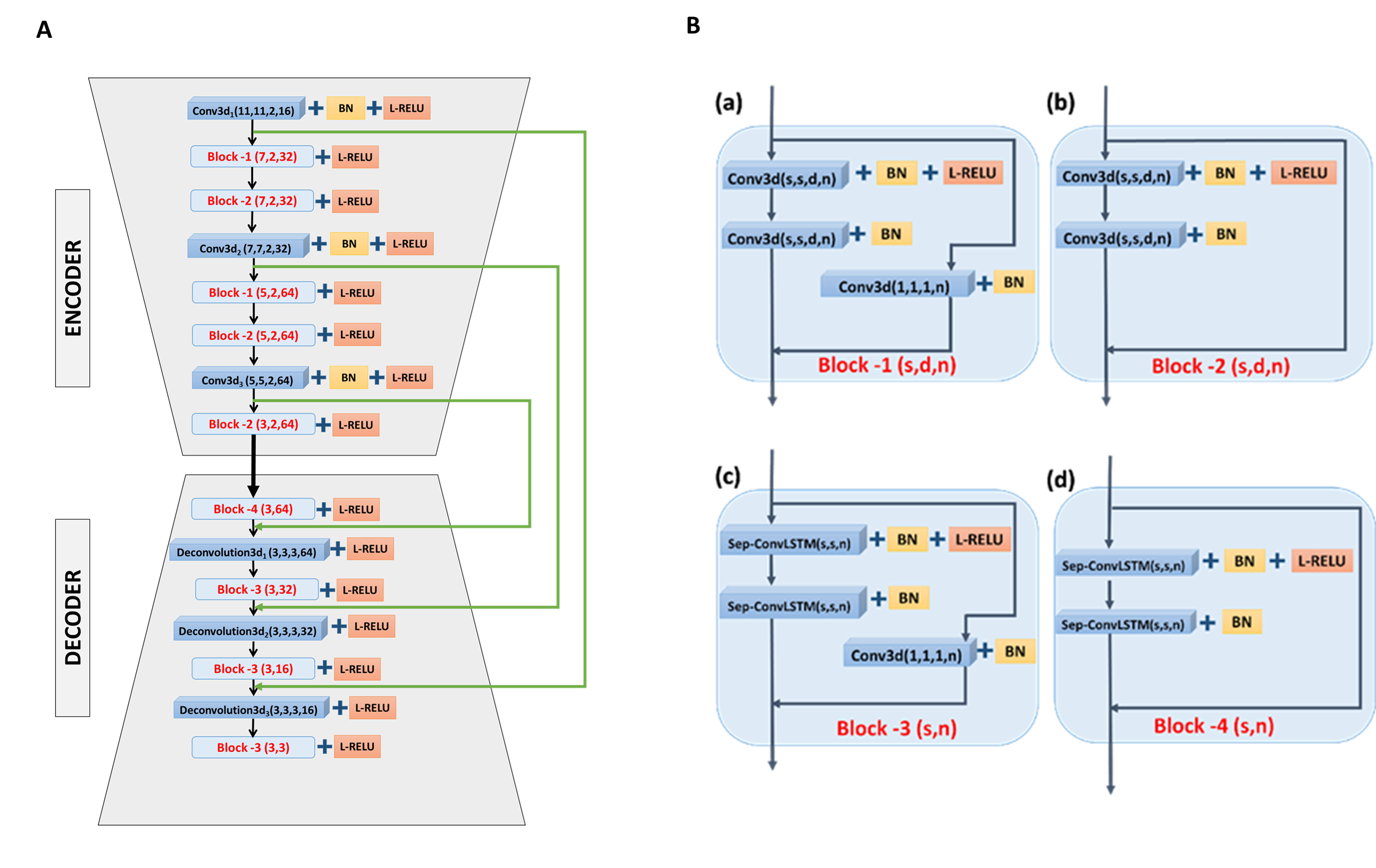}
\end{center}
  \caption{A- Overview of the proposed encoder/decoder network used in our approach to predict the next $N$ video frames (the future) using the first $N$ video frames (the past) as input. B- The 4 different residual blocks used in the architecture in A. (a) and (b) are the residual blocks used in the encoder. (c) and (d) are used in the decoder.}
\label{fig:onecol}
\end{figure*}
The future frames prediction component of the model was an unsupervised encoder/decoder that predicted future frames in a video sequence. $N$ consecutive video frames were input into the model, and the model predicted $N$ frames into the future.  
Figure \ref{fig:onecol}A is a visual representation of the encoder/decoder architecture. The encoder mapped the input sequence of frames into a low-dimensional feature space with different resolutions. The decoder mapped the low-dimensional representation space of the input frames to an output image space. 
We define the input as a sequence $x = \{x_1 , \dots , x_N \}$ where $x_i$ refers to the frame of index $i$ in the input. The encoder yielded dense representations $z = \{z_1, ..., z_N\}$ from the input. The decoder then output the following frames, denoted as $y' = \{y'_{N+1}, \dots, y'_{2N} \}$, as a prediction of the ground-truth frames $y = \{y_{N+1}, \dots, y_{2N} \}$. 

Next, we will detail the precise structure of the encoder and decoder.  

\textbf{Encoder:} The encoder was a spatio-temporal neural network composed of three-dimensional convolutional layers. 3D convolutions modeled both the spatial and sequential relationships of temporal connections across frames. $N$ RGB frames were the encoder input. The size of input is $3 \times N \times H \times W$.  The temporal duration of the outputted feature maps matched the input images. 

The main components of the encoder were Block-1, Block-2 and 3D convolutions (conv3d) layers (as shown in Figure \ref{fig:onecol}). Each convolutional operation was followed by batch normalization \cite{ioffe2015batch} and a Leaky-ReLU activation function. The residual blocks (Block-1 and Block-2 in Figure \ref{fig:onecol}B) consisted of two 3D convolutions with no stride. Residual connections addition operations were identity shortcuts for Block-2 and $1 \times 1 \times 1$ 3D convolution operations matched the input and output dimensions in Block-1. Inputs were downsampled in $conv3d_1$, $conv3d_2$, and $conv3d_3$ with a stride of 2. The filters of the last Block-2 of our encoder were time-dilated filters which captured the temporal dependency with various resolutions for the input sequences. 

\textbf{Decoder:} 
The decoder was composed of convLSTMs layers interspersed with up-sampling layers. 
The main components of the decoder architecture are Block-3 and Block-4 shown in Figure \ref{fig:onecol}B. The blocks were composed of two consecutive depth-wise separable convLSTMs with a residual connection connecting the input to the output. The residual connection in Block-4 was a simple identity shortcut, while Block-3 was $1 \times 1 \times 1$ 3D convolution operations matching the input and output dimensions. For the up-sampling layers , we used deconvolution \cite{dumoulin2016guide} as it uses learnable parameters rather than predefined up-sampling.

\textbf{Encoder/Decoder Connections:} 
Lateral skip connections crossed from same-sized parts in the encoder to the decoder (the green lines shown in Figure \ref{fig:onecol}). The lateral connections increased the amount of frames details, resulting in increased detail in the predicted frames.

\subsubsection{Pedestrian Action Prediction Component}\label{classification}
The second stage of the model consisted of a fine-tuned early action recognition network, the 'Temporal 3D ConvNets' (T3D) \cite{diba2017temporal}. This stage predicted if the pedestrians would cross the street in the scene. 
$N$ predicted frames, generated from the encoder/decoder, were input into the network. The last classification layer of the T3D network was replaced with a fully connected layer, which produced a single output followed by sigmoidal activation. The component was trained with binary-cross entropy loss. 

\subsubsection{End-to-end Loss}\label{loss}

The model had two main tasks that could be used for training: the future frames prediction loss and the pedestrian crossing loss. The full model was trained end-to-end with a multi-task learning objective:
\begin{equation}
    L_{recog} = \lambda L_{pred} + L_{ce}(Y,\hat{Y}),
\label{crossentropy}
\end{equation}
where $L_{ce}$ was the cross-entropy loss for crossing action classification, $\hat{Y}$ and $Y$ were high-level predictions and corresponding ground truth classes. The weight factor of the loss was $\lambda$. $L_{pred}$ was the future frame prediction loss, the pixel-wise loss between the pixels of the $N$ predicted frames and the $N$ ground-truth frames, defined as:
\begin{equation}
    L_{pred} = \frac{1}{P}( \sum^{2N}_{t=N+1}\sum^{P}_{i=1}(y_{t,i} - y'_{t,i})^2 + \sum^{2N}_{t=N+1}\sum^{P}_{i=1}|y_{t,i} - y'_{t,i}|)
\end{equation}
where $P = H \times W $, the number of pixels per frame. We use a combination of $l_1$  and $l_2$ norm losses for regularization purposes.

\subsection{Experiments}\label{experiments}

In this subsection, we describe the dataset that was used to evaluate our model (\ref{data}), the model selection search details (\ref{model_search}), and the experimental details of our experiments (\ref{setup}).

\subsubsection{Dataset}\label{data}

The Joint Attention for Autonomous Driving (JAAD) dataset was used for training and evaluation across all experiments \cite{rasouli2017they}. JAAD was designed to study the behavior of traffic participants, with a focus on pedestrian and the driver behavior. Approximately $240$ hours of driving videos were collected in several locations using two vehicles equipped with wide-angle video cameras. Cameras were mounted in the center of the cars' windshields, below the rear view mirror. The dataset consists of 346 high-resolution video clips that focus on pedestrian and driver behaviors at the point of crossing. Most of the data was collected in urban areas, but some clips were filmed in rural areas. 

We picked JAAD because it contained a variety of interactions and complex situations that may impact the behavior of the traffic participants. Complex Interactions included pedestrians crossing in groups or individually, partially occluded pedestrians, and pedestrians walking parallel to the street. Complex situations included interactions with other drivers in parking lots or uncontrolled intersections. A variety of difficult conditions are represented in the dataset, such as weather, light conditions, and day/night conditions. 

\subsubsection{Model Search}\label{model_search}
We conducted a large-scale model search to identify the best performing encoder/decoder model for our future video frames prediction component. The search involved three steps, which are further detailed below. First, we selected a potential architecture design. Second, we trained 38 variations of the architecture with random sets of hyperparameters. Finally, the best architecture/hyperparameters combination was identified by lowest error on average future frames prediction, across $N$ frames.

The architecture and hyperparameters presented in Figure \ref{fig:onecol} represent the combination with the highest performance.

\textbf{Architecture Design:} The main encoder/decoder components were experimentally manipulated in order to test multiple architectural designs. The number of layers, the order of the layers, and the number of channels within layers were all varied. Across all variations, the encoder output remained unchanged because the spatial dimension of the input was consistently downsampled by 8. In the decoder, the convLSTM block(s)-deconvolution pattern was consistently used.

\textbf{Hyperparameters Selection:} For each selected architecture, 38 hyperparameter settings were randomly sampled \cite{10.1093/bioinformatics/btz339}. Each parameter setting was evaluated using its average pixel-wise prediction $l_1$ error on the validation set. Details of the hyperparameter search spaces are summarized in Table \ref{table_hyperparameters}.

\begin{table}
\begin{center}
\begin{tabular}{|l|c|c|}
\hline
Calibration parameters  & search space \\
\hline\hline
Spatial filter size of 3D Convs  & [3,5,7,11]\\
Temporal dilation rate  & [1,2,3,4] \\
Spatial filter size of sep-ConvLSTMs  & [3,5,7] \\
Temporal filter size of 3D Convs & [2,3,4] \\
Temporal filter size of sep-ConvLSTMs & [2,3,4] \\
\hline
\end{tabular}
\end{center}
\caption{encoder/decoder network hyperparameters and search space. Note: temporal dilation rate is implemented only in the last Block of the encoder.}
\label{table_hyperparameters}
\end{table}

\subsubsection{Experimental Setup}\label{setup}

We used the same training, validation, and test clips presented in \cite{gujjar2019classifying}, which allowed us to directly compare our performance. $60\%$ of the data was set aside for training, $10\%$ for validation and $30\%$ for testing. Clips were divided into $2N$-frame videos with a temporal stride of $1$. The frames were resized to $128 \times 208$, with $N = 16$. Thus, the model input was $3 \times 16 \times 128 \times 208$. 

The future frames prediction component was pre-trained on JAAD dataset and the T3D network was pre-trained on UCF101 dataset \cite{soomro2012dataset}. The model was fine-tuned on the end-to-end loss, as detailed in Section \ref{loss}, with the Adagrad \cite{duchi2011adaptive} optimizer. The weight factor $\lambda$ was set to $0.5$. The end-to-end model was fine-tuned for $30$ epochs with a learning rate of $10^{-4}$. Our end-to-end model was fine-tuned with the same JAAD splits described in
\cite{gujjar2019classifying,rasouli2017they} for comparison. The test set is composed of $1257$ $16$-frames videos of which $474$ are labelled as crossing and $783$ as not crossing. 

All experiments were run on an Ubuntu server with a TITAN X GPU with $12$ GB of memory. 

\section{Results}\label{results}
We evaluated our model on two predictive tasks: future frames and future pedestrian crossing. In this section, we present a complete description of our experiments and subsequent results. In summary, our model outperformed state-of-the-art models in predicting future pedestrian crossing action on JAAD dataset, yielding $86.7$ Average Precision (AP) as compared to $81.14$ AP achieved by the best performing approach \cite{gujjar2019classifying}. 
The source code and trained models will be made available to the public.

\subsection{Future Frames Prediction}\label{results_FFP}

Future Frames Prediction is the problem of accurately generating future frames given a set of consecutive previous frames. We quantitatively compared our model's performance on future frames prediction against other state-of-the-art methods (Section \ref{quant}), detailed consistent architectural trends that we identified from our model search (Section \ref{trends}), conducted an Ablation Study (Section \ref{ablation}), and performed a qualitative analysis on our model (Section \ref{qualt}). 

\subsubsection{Quantitative Analysis}\label{quant}

Most state-of-the-art methods for future frame prediction have published results on datasets that have static cameras and identical backgrounds, such as the Moving MNIST dataset \cite{srivastava2015unsupervised} and the KTH action dataset \cite{schuldt2004recognizing}. In this work, we evaluated and compared our model to state-of-the-art models on the JAAD dataset, which consists of complex, real-world interactions and variability that we can expect autonomous vehicles to encounter regularly (we detail the complexities of the dataset in Section \ref{data}). We limited our quantitative comparison to state-of-the-art methods which have publicly available code, including PredRNN++ \cite{wang2018predrnn++} and PredNet \cite{lotter_deep_2017}, and Res-EnDec \cite{gujjar2019classifying}, who trained and tested on the same data split we used (see Section \ref{setup}). 

PredRNN++ and PredNet were originally designed to predict one frame ahead, but they were modified to predict multiple future frames by treating their predicted frame as input and recursively iterating. We trained both models from scratch on JAAD dataset on the same train/validation/test described in our methods.  

Res-EnDec \cite{gujjar2019classifying} is, to the best of our knowledge, the best performing model with published results on JAAD dataset. The code for this model was not released, so we directly compared our results to the results reported in \cite{gujjar2019classifying}. 

\begin{table}
\begin{center}
\begin{tabular}{|l|c|}
\hline
Model & \textit{$l_1$} loss  ($ \times 10^{-1}$)  \\
\hline\hline
Res-EnDec \cite{gujjar2019classifying} & $ 1.37  \pm  0.37$\\
PredRNN++ \cite{wang2018predrnn++} &  $1.61  \pm  0.35$\\
PredNet \cite{lotter_deep_2017} & $1.30 \pm 0.41$\\
Ours & $\textbf{1.12} \pm \textbf{0.32}$ \\

\hline
\end{tabular}
\end{center}
\caption{Comparison of our model with state-of-the-art methods on JAAD dataset. We report pixel-wise prediction $l_1$ error averaged over the $16$ predicted frames.}
\label{table_comparison_l1}
\end{table}

In Table\ref{table_comparison_l1}, we present the average performance of each model on the future frame prediction task across 16 time steps. Our model has the lowest error among the models we tested. 

\begin{figure}[ht]
\begin{center}
   \includegraphics[scale = 0.3]{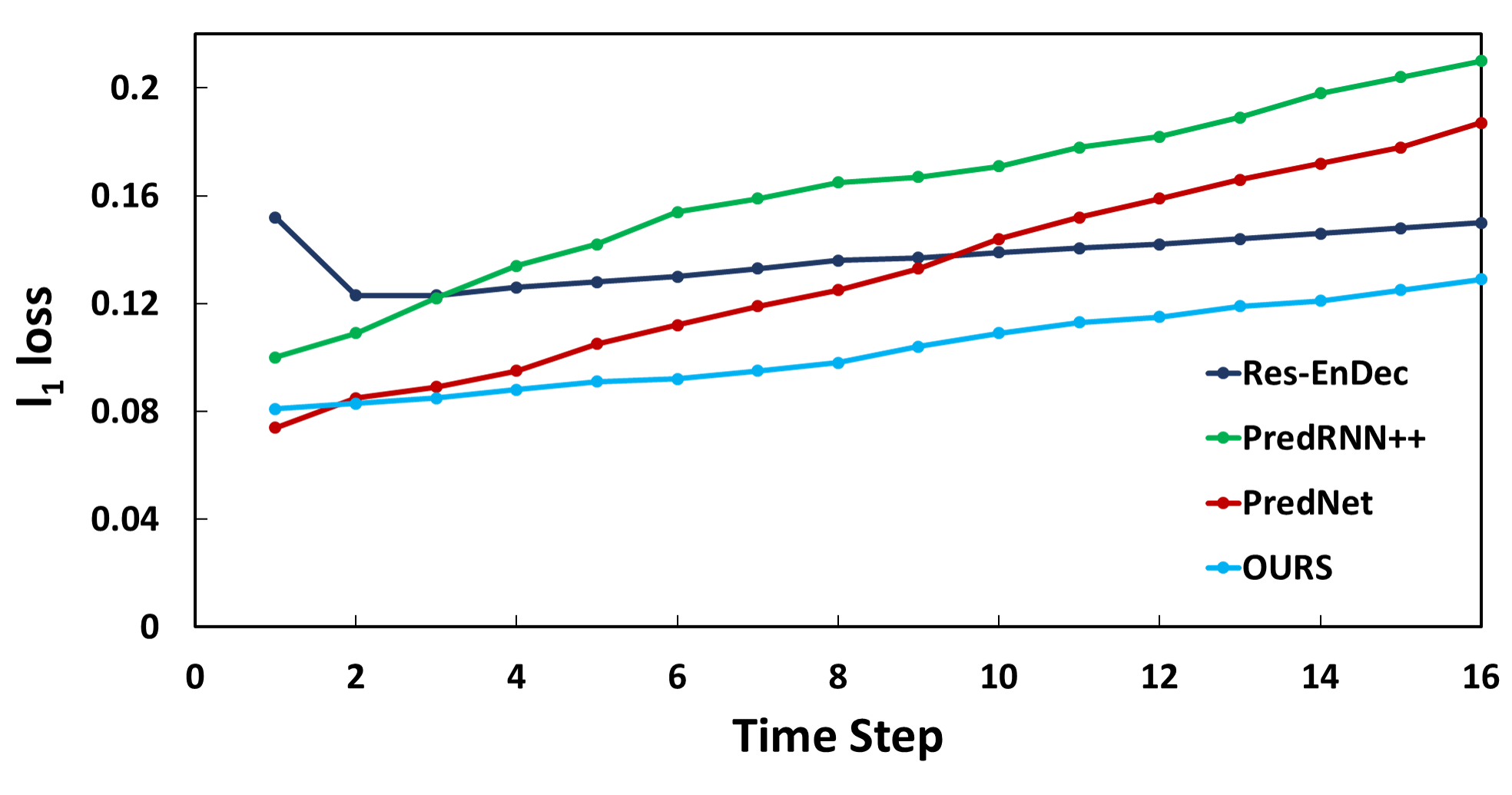}
\end{center}
  \caption{
  Comparison of our model with state-of-the-art models on JAAD dataset. We report frame-wise $l_1$ loss of generated sequences on the JAAD test set. All models are trained from scratch on JAAD training set.}
\label{fig:quant}
\end{figure}

In Figure \ref{fig:quant}, we plotted the $l_1$ loss of each model between the predicted and the ground truth frames across multiple time steps, up to $16$ frames into the future. This allowed us to evaluate each model's relative consistency across the predicted frames. The quality of the predicted frames degraded over time for all models. Res-EnDec model had a slight variation in this trend; the error was higher for time step 1, explained in \cite{gujjar2019classifying} as a result caused by the reverse ordering of their inputs.   

Our model outperformed the state-of-the-art methods for all time steps except the initial time step, where PredNet produced slightly better performance. 
PredRNN++ and PredNet produced reasonably accurate short term predictions, however, they broke down when extrapolating further into the future. Compared with our model and \cite{gujjar2019classifying}, their errors increased considerably over time. This is expected, since both PredRNN++ and PredNet are not explicitly designed and optimized to predict multiple future frames. The predicted frames unavoidably had different statistics than the natural images the models were optimized for \cite{bengio2015scheduled}. Given that, it is unsurprising that our model and \cite{gujjar2019classifying} have better performance for long term predictions.

\subsubsection{Architectural Trends Identified from Model Search}\label{trends}

The performance of our selected future frames prediction component, shown in Figure \ref{fig:quant}, resulted from the model selection criteria explained in Section \ref{model_search}. Through the model search process, we discovered several interesting and consistent architectural trends and their relationship to next frame prediction performance and speed. We report these trends here, so they can inform future model searches related to this topic. 

We found that kernels with decreasing sizes in the spatial dimension ($11\times11 \xrightarrow{}
7\times7 \xrightarrow{} 5\times5 \xrightarrow{} 3\times3$) and constant size in the time dimension exhibited higher performance across temporal variations, while still capturing the details from the scene.
We also noted that ascendant dilation rates in the temporal dimensions at the end of the encoder network ($1 \xrightarrow{} 2 \xrightarrow{} 4$) performed best, and enabled our encoder network to have larger temporal receptive field without the need to add more layers.

\subsubsection{Ablation Study}\label{ablation}

In Table \ref{table_ablation} we present the results of an ablation study investigating how 7 architectural variants (V-1 - V-7) affected our model's performance and speed. 

\begin{table}
\begin{center}
\begin{tabular}{|l|c|c|c|}
\hline
Model & \textit{$l_1$} loss  ($ \times 10^{-1}$)  & SSIM & Time (ms)\\
\hline\hline
Ours & $1.12 \pm 0.32 $ & $0.924$ & $88.74$\\
\hline
V-1 (Reg-convLSTM) & $1.13 \pm 0.39$ & $0.920$& $100.36$\\
V-2 (Spatial-convLSTM) & $1.11 \pm 0.31$ & $0.929$ & $109.84$\\
V-3 (Depth-convLSTM) & $1.16 \pm 0.35$ & $0.911$& $80.69$\\
V-4 (w/o laterals)& $1.26 \pm 0.37 $& $0.883$& $86.69$\\
V-5 (w/o residuals)& $1.23 \pm 0.34$ & $0.894$& $85.31$\\
V-6 (Undilated) & $1.20 \pm 0.29$ & $0.905$& $87.43$\\
V-7 (w/o Deconv) & $1.18 \pm 0.43$ & $0.909$& $87.25$\\

\hline
\end{tabular}

\end{center}
\caption{Ablation study of different model variants on the JAAD dataset. We report the average pixel-wise prediction $l_1$ error, the per-frame structural similarity index measure (SSIM) \cite{wang2004image} averaged over the $16$ predicted frames and the running time.}
\label{table_ablation}
\end{table}

In the first three architecture variants, we experimented with convLSTM variants in the decoder. Our model used a depth-wise separable convLSTM. Variant one (V-1) consisted of standard convLSTMs, V-2 was spatially separable convLSTMs, and V-3 was depth-wise convLSTMs. SSIM performance across all three variants was similar, but V-2 (used spatial convLSTM layers) performed slightly better than the other variants. In the end, we opted for depth-wise separable convLSTMs because our network ran $21\,ms$ faster than V-2, with very similar performance.

The remaining variants removed or replaced network components, highlighting the importance of each component. In V-4, we removed the lateral connections, stymieing pixel information in the deconvolution. 
In V-5, residual connections were removed, underscoring the importance of this feature. 
In V-6, the temporal dilation was set to 1, limiting the model's temporal information. 
Finally, in V-7, we replaced deconvolution with interpolation. We suspect deconvolution can reconstruct the shape and boundaries more accurately than interpolation. 

\subsubsection{Qualitative Analysis}\label{qualt}
\begin{figure}[ht]
   \includegraphics[scale = 0.52]{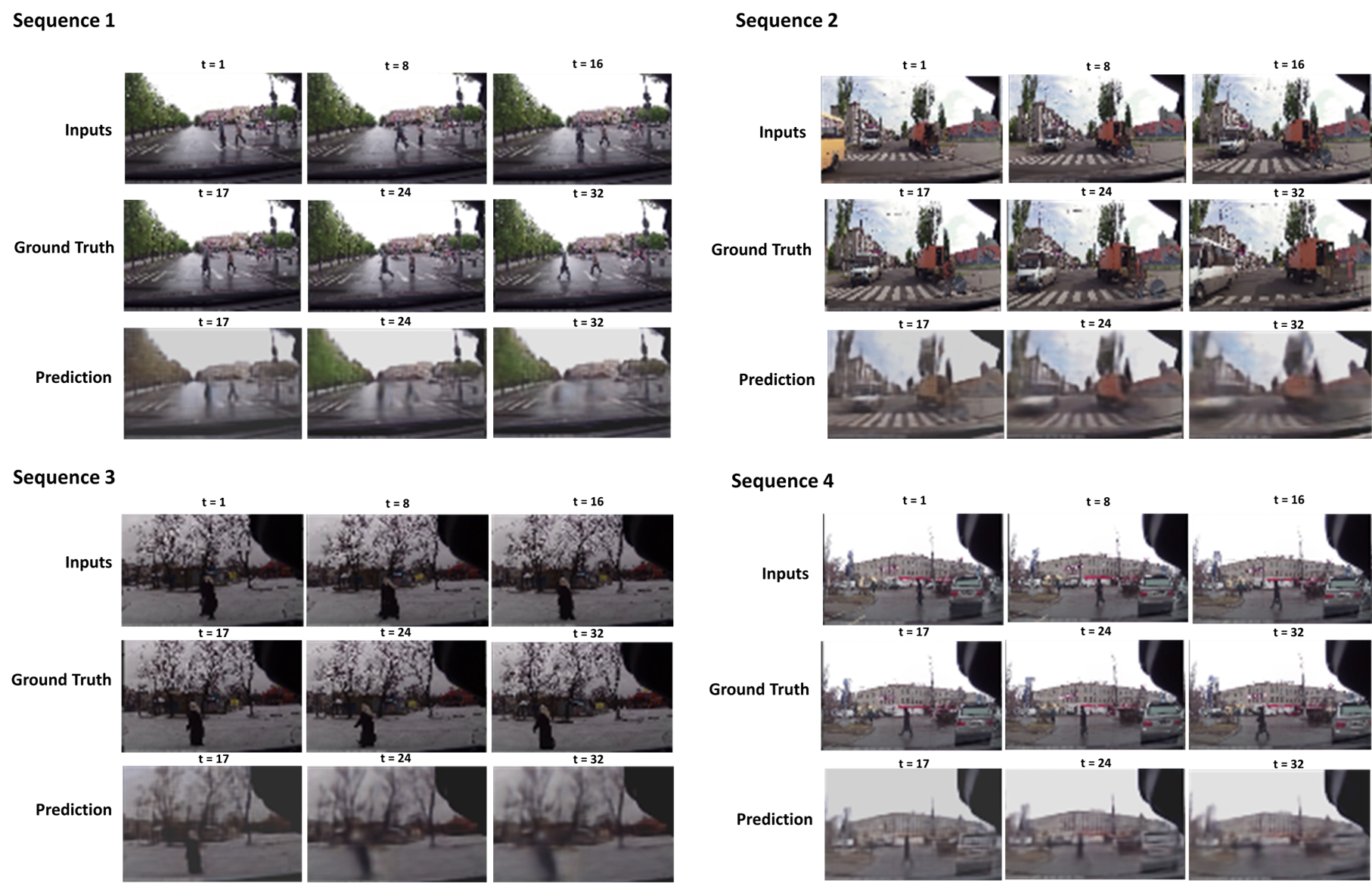}
  \caption{ Future frames prediction examples on JAAD dataset. We predict $16$ frames into
the future by observing $16$ frames.  }
\label{fig:qual}
\end{figure}
Sample predictions for our model on the JAAD dataset are shown in Figure \ref{fig:qual}. The
model is able to make accurate predictions in a wide range of scenarios. In sequence 1 and sequence 3
of Figure \ref{fig:qual}, pedestrians are crossing the street in two different weather conditions (rain/snow), and the model successfully predicts their positions moving forward from frame to frame. Similarly in Sequence 2, a car is passing in the opposite direction, and the model, while not perfect, is able to predict its trajectory, as well as the position of the zebra stripes of the crossing area and other stationary objects as the driver moves forward. 
Sequence 4 illustrates that our model can distinguish between moving pedestrians and standing pedestrians as it accurately predicts the movement of the pedestrian who is crossing the street while the second pedestrian in the left-hand side of the scene was still standing in their same initial position.

\subsection{Pedestrian Future Crossing Action Prediction}

\begin{table}
\begin{center}
\begin{tabular}{|l|c|c|}
\hline
Model & AP & Time (ms)  \\
\hline\hline
C3D & $84.9$  & $\textbf{93.16}$\\
Inception-3D &  $82.2$ & $104.68$  \\
ResNet3D-50 & $79.6$  & $99.45$\\
T3D & $\textbf{86.7}$  & $102.48$\\

\hline
\end{tabular}
\end{center}
\caption{ Evaluation results of future crossing action prediction with different action recognition networks as extension to our future frames prediction component. We report Average Precision (AP) and the running time. All action recognition networks are pre-trained on UCF101 dataset and trained using same multi-task training strategy.}
\label{table_3dconvnets}
\end{table}

\begin{table}
\begin{center}
\begin{tabular}{|l|c|}
\hline
Model & AP   \\
\hline\hline
Action \cite{rasouli2017they} & $39.24  \pm 16.23$  \\
Action + Context \cite{rasouli2017they} &  $62.73  \pm 13.16$  \\
Res-EnDec \cite{gujjar2019classifying} & $81.14$ \\
Ours ( Separate training ) & $85.8$  \\
Ours ( Joint training ) & $\textbf{86.7}$  \\

\hline
\end{tabular}
\end{center}
\caption{Comparison of average precision (AP) of our model (with two different training strategies) with state-of-the-art methods for predicting future crossing action on JAAD dataset}
\label{table_comparison}
\end{table}

As described in Section \ref{classification}, we extended our future frames prediction component with the T3D network to classify a predicted future scene as pedestrian crossing/not crossing. 
We compared T3D with some variants where we replace T3D network with other 3D ConvNets networks for action recognition including C3D \cite{tran2014c3d} and 3D ConvNets based on Inception \cite{szegedy2015going} and ResNet-34 \cite{he2016deep}. To make the comparison fair, all action recognition networks were pre-trained on the UCF101 dataset and then end-to-end trained to minimize the multi-task learning objective function shown in equation \ref{crossentropy}. Table \ref{table_3dconvnets} compares average precision (AP) and running time for predicting future crossing action using only the generated $16$ frames. Running time in Table \ref{table_3dconvnets} corresponds to our model's $16$-frame input and $16$-frame prediction, as well as the prediction of future crossing action. T3D network outperformed other 3D convNets, likely because it modeled variable temporal 3D convolution kernel depths over shorter and longer time ranges. 

Using JAAD dataset, predicting $16$ future frames at a frame rate of $30 fps$ corresponds to looking ahead $533 ms$ in time. Any advantage gained from this is reduced by the running time of the model. The running time for our model is $102.48 ms$ which provides a reliable maximum look-ahead time of $430 ms$. This allows $81\%$ of the time before the occurrence of the final predicted frame to be utilized for defensive reaction.

We also compared AP scores of our model with the results presented by Rasouli et al. in \cite{rasouli2017they} and Gujjar et al. in \cite{gujjar2019classifying} in Table \ref{table_comparison}. Our model outperformed Action \cite{rasouli2017they}, Action + Context \cite{rasouli2017they} and Res-EnDec \cite{gujjar2019classifying} by about $47.5 \%$, $24 \%$ and $5.6 \%$, respectively. Our model and Res-EnDec model outperformed models in \cite{rasouli2017they} with more than $18 \%$, which reflects the effectiveness of predicting future video frames and using those frames to predict future crossing action. Using same action recognition network (C3D) as in \cite{gujjar2019classifying}, our model achieves $84.9$ AP (Table \ref{table_3dconvnets}) which is $3.8 \%$ higher than Res-EnDec model. Likely, some of our improvements can be attributed to the increased quality of our generated future frames. The remaining improvements stem from the multi-task training strategy we employ --- our model's AP rose $0.9 \%$ when we implemented this approach, in agreement with previous findings \cite{ruder2017overview}.

\section{Conclusion}

In this paper, we introduced an end-to-end future-prediction model that uses previous video frames to predict the pedestrian's future crossing action. Our multi-task model is composed of two stages. The first stage consists in processing $N$ input video frames using an encoder/decoder network to predict $N$ future frames. The second stage utilizes the predicted frames to classify the pedestrian's future actions. Our end-to-end model predicts a future pedestrian crossing action with an effective look-ahead of $430\,ms$ on JAAD dataset. 

Our end-to-end model achieves state-of-the-art performance in model speed ($102.48\,ms$), predicting future frames ($0.924\,SSIM$), and predicting pedestrians crossing ($86.7\,AP$) on JAAD dataset. Our ablation study demonstrates the effectiveness of the different model's components on the model performance. Further quantitative and qualitative experiments have shown the ability of our proposed model across the various weather conditions, locations and other variables included in the JAAD dataset.

While this is just one step in replicating human-like behavior for self-driving cars, the safety of pedestrians is an essential first step to tackle, and our work has promising implications for the future of replicating human-like behavior in the context of fully autonomous driving.

\bibliographystyle{unsrt}  
\bibliography{references}  





\end{document}